\pdfoutput=1
\documentclass{article}

\usepackage[preprint]{preprint}
\usepackage{amsmath}
\usepackage{amssymb}
\usepackage{graphicx}
\usepackage{booktabs}
\usepackage[hidelinks]{hyperref}
\usepackage{algorithm}
\usepackage{algorithmic}
\usepackage{microtype}  

\preprinttitlerunning{Surprisal-Guided Selection}

\begin{document}

\twocolumn[
\preprinttitle{Surprisal-Guided Selection: Compute-Optimal Test-Time Strategies\\for Execution-Grounded Code Generation}

\preprintsetsymbol{equal}{*}

\begin{preprintauthorlist}
\preprintauthor{Jarrod Barnes}{ai}
\end{preprintauthorlist}

\preprintaffiliation{ai}{Arc Intelligence}

\preprintcorrespondingauthor{Jarrod Barnes}{jarrod@arc.computer}

\preprintkeywords{test-time compute, surprisal-guided selection, GPU kernel optimization, KernelBench, Best-of-N sampling, model calibration, verifiable execution}

\vskip 0.3in
]

\printAffiliationsAndNotice{}

\begin{abstract}
Test-time training (TTT) adapts language models through gradient-based updates at inference. But is adaptation the right strategy? We study compute-optimal test-time strategies for verifiable execution-grounded (VEG) tasks, domains like GPU kernel optimization where a deterministic evaluator provides dense, continuous reward signals.

Using KernelBench as our testbed and a 120B-parameter model (GPT-OSS-120B with LoRA adaptation), we find that \textbf{search outperforms minimal adaptation (1-5 gradient steps)}: Best-of-N sampling achieves 90\% task success (18/20 tasks) at $K\!=\!64$ across the full KernelBench L1 eval set while TTT's best checkpoint reaches only 30.6\% (3-seed mean), with TTT's ``equivalent $K$'' falling below 1, worse than single-sample inference. The failure mode is over-sharpening: gradient updates collapse diversity toward mediocre solutions rather than discovering optimal ones.

Our main contribution is \textbf{surprisal-guided selection}: selecting the \emph{highest-surprisal} (lowest-confidence) correct sample yields 80\% success vs.\ 50\% for most-confident selection, a 30\% improvement. Extending to surprisal-guided-top3 matches oracle performance at 100\%. This zero-cost strategy, validated through length-controlled analysis, recovers oracle performance.

For dense-reward VEG tasks, compute should be allocated to sample diversity and intelligent selection rather than gradient adaptation. The surprisal-guided selection principle may generalize to other execution-grounded domains where optimal solutions occupy the distribution tail.
\end{abstract}

\section{Introduction}

This paper studies compute-optimal test-time strategies for verifiable execution-grounded (VEG) tasks, domains where a deterministic evaluator provides ground-truth feedback on model outputs. GPU kernel optimization exemplifies VEG: KernelBench~\cite{kernelbench2025} evaluates 250 PyTorch ML workloads on both functional correctness and runtime speedup, with the CUDA compiler and hardware providing an unambiguous, continuous reward signal. The defining characteristic is that the evaluator provides ground-truth feedback; no human labeler or AI teacher is needed to judge output quality.

\begin{figure}[t]
\centering
\includegraphics[width=\columnwidth]{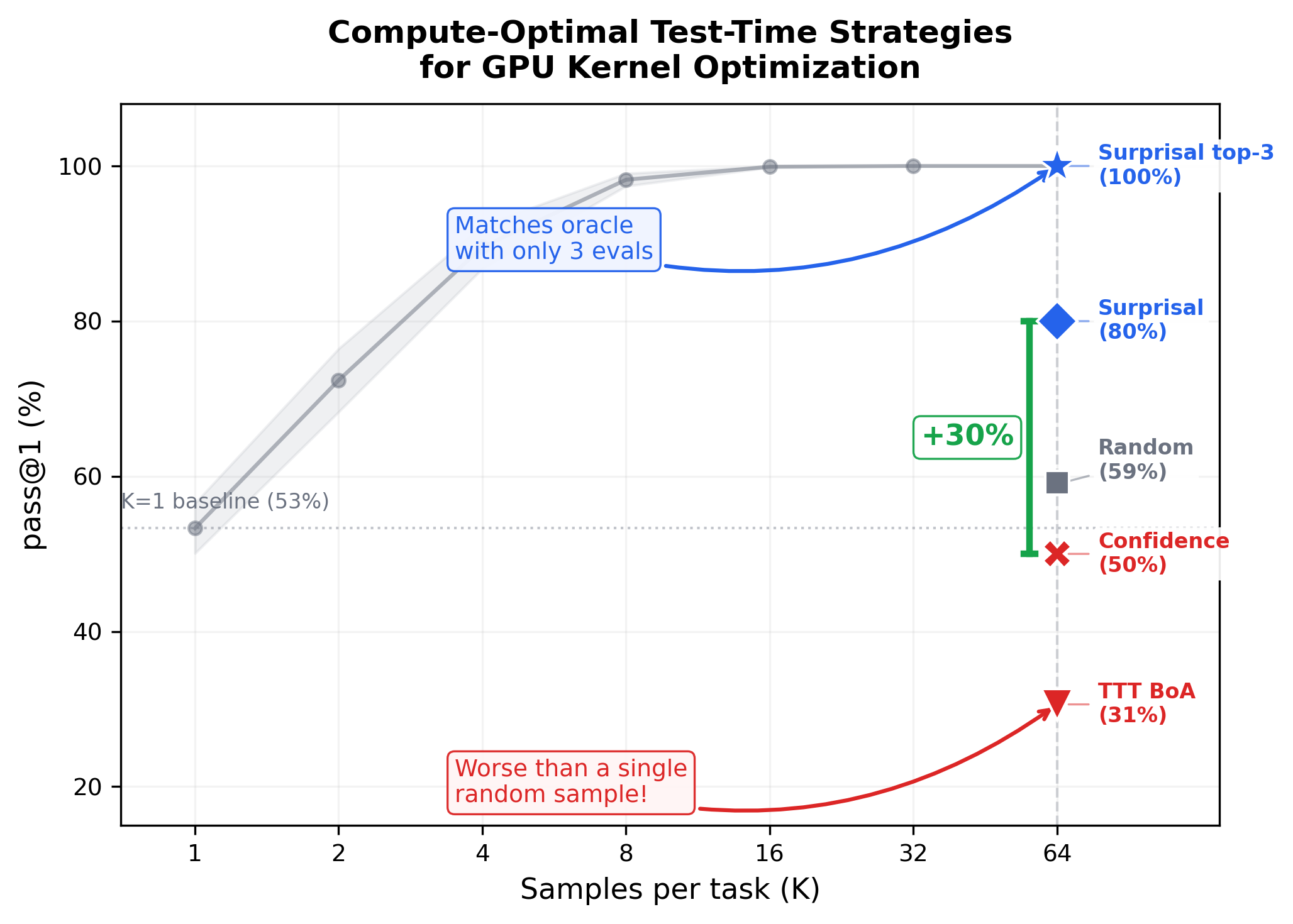}
\caption{\textbf{Test-time strategy comparison.} Best-of-N scaling (gray) saturates at $K\!=\!16$. At $K\!=\!64$, TTT (31\%, red) is \textbf{2$\times$ worse than random selection} (59\%); surprisal-guided (blue) matches oracle. The +30\% bracket: confidence (50\%) vs.\ surprisal (80\%).}
\label{fig:teaser}
\end{figure}

\textbf{Why VEG tasks are the ideal testbed.} Unlike binary pass/fail benchmarks, KernelBench provides \emph{continuous} speedup signals (0x to 10x+). This density enables us to detect subtle performance regressions during adaptation that binary rewards would mask. When TTT over-sharpens, we observe the decline in a continuous metric; papers with sparse rewards may miss this entirely.

Recent work on test-time training (TTT) has shown impressive results through extended gradient-based adaptation. TTT-Discover~\cite{tttdiscover2026} reports costs of ``a few hundred dollars per problem'' using $\sim$50 adaptation steps on discovery tasks. This raises a fundamental question: \textbf{is adaptation the right strategy for dense-reward VEG tasks, or does simple search suffice?}

We answer this question through controlled experiments comparing Best-of-N sampling against batch test-time training under matched compute budgets. Using GPT-OSS-120B (a 120B-parameter frontier model) with LoRA adaptation, we evaluate all 20 KernelBench L1 eval tasks. The results are decisive. Best-of-N at $K\!=\!64$ achieves 90\% task success (18/20 tasks, finding at least one fast correct kernel per task) while TTT's best checkpoint (Best-of-Adaptation) reaches only 30.6\% (3-seed mean). Computing TTT's ``equivalent $K$'' (the Best-of-N budget needed to match TTT performance) yields $K < 1$, meaning TTT underperforms \emph{single-sample inference ($K\!=\!1$)}.

The failure mode is \textbf{over-sharpening}: gradient updates collapse the policy toward mediocre solutions that happened to succeed early, destroying the diversity needed to find optimal kernels in the distribution tail. Ji et al.\ \cite{scalable_power_sampling2026} predict that RL gains arise from distribution sharpening rather than discovering new strategies; our failure mode confirms this.

\textbf{Our main contribution is surprisal-guided selection.} Probing the relationship between model confidence (log-probability) and kernel quality reveals a surprising inverse correlation: the model is \emph{least} confident about its best solutions. We operationalize this as \textbf{surprisal-guided selection}: selecting the \emph{highest-surprisal} (lowest log-probability) correct sample. This achieves 80\% success (fast and correct) versus 50\% for confidence-guided selection, a 30\% improvement with zero additional compute. Extending to \textbf{surprisal-guided-top3} (evaluating the 3 highest-surprisal correct samples and selecting the fastest) matches oracle performance at 100\%.

Three contributions emerge from our experiments (Figure~\ref{fig:teaser}):

\vspace{0.5em}
\begin{enumerate}
\item \textbf{Search outperforms minimal adaptation} (1-5 GRPO steps) \textbf{for dense-reward VEG tasks.} Best-of-N scaling saturates at $K\!=\!16$ (99.9\% success on 5-task subsets; 90\% on the full 20-task L1 eval), while TTT equivalent $K < 1$. Practitioners should invest in sample diversity, not gradient updates.

\item \textbf{Surprisal-guided selection recovers oracle performance.} Selecting from the high-surprisal tail (solutions the model ``didn't expect to find'') provides a practical, zero-cost selection strategy.

\item \textbf{Mechanistic explanation for TTT failure.} Over-sharpening destroys diversity, confirmed by direct correlation probing. The optimum for kernel optimization lies in the distribution tail; gradient updates collapse toward the mode, missing the tail entirely.
\end{enumerate}

For execution-grounded domains with dense rewards, compute should be allocated to sample diversity and intelligent selection rather than gradient adaptation. The surprisal-guided selection principle (that the model's best solutions occupy high-surprisal regions) may generalize to other VEG domains where rare, high-quality solutions exist in low-probability regions of the model's distribution.

\begin{figure*}[t]
\centering
\includegraphics[width=0.85\textwidth]{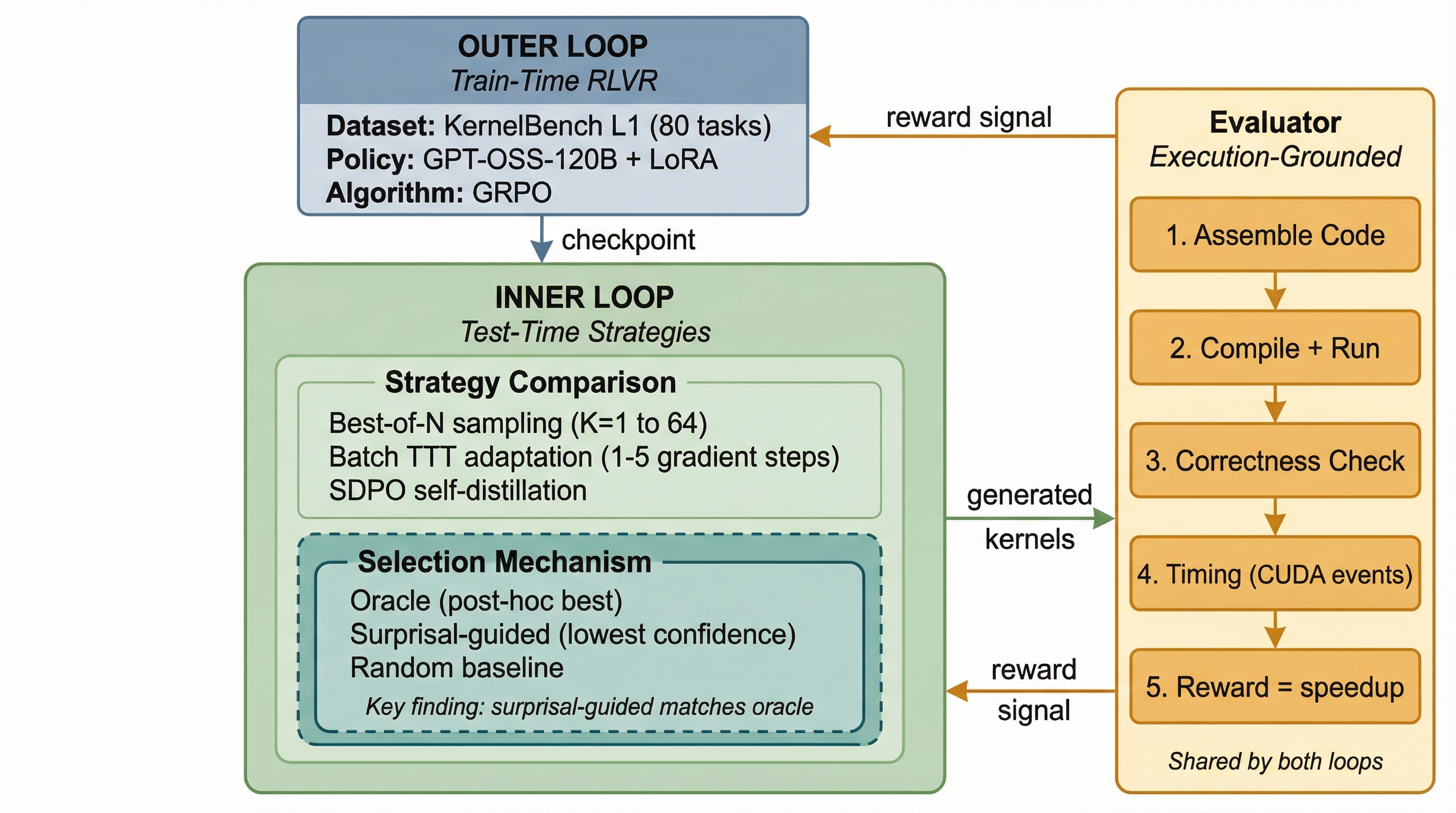}
\caption{\textbf{Dual-loop architecture.} The outer loop (blue) trains a base policy via reinforcement learning from verifiable rewards (RLVR) on 80 KernelBench tasks. The inner loop (green) compares test-time strategies under matched compute. The \textbf{selection mechanism} (dashed box) determines how to choose among correct samples. Both loops share the same evaluator (orange).}
\label{fig:architecture}
\end{figure*}

\section{Related Work}

\textbf{Test-Time Training vs.\ Search.} TTT-Discover~\cite{tttdiscover2026} demonstrates impressive results using $\sim$50 adaptation steps on discovery tasks, reporting costs of ``a few hundred dollars per problem.'' We find different dynamics for dense-reward VEG tasks: search outperforms 1-5 step adaptation, with TTT's best checkpoint underperforming even random sampling. The difference likely stems from reward density: TTT-Discover targets sparse-reward scientific discovery where extended exploration may shift the distribution. Kernel optimization has dense rewards; optimal solutions already exist in the base distribution's tail (see Section~5.6 for detailed comparison).

\textbf{Distribution Sharpening and Over-Fitting.} Scalable Power Sampling~\cite{scalable_power_sampling2026} argues that RL gains arise from distribution sharpening rather than discovering qualitatively new strategies. Our TTT failure mode provides empirical evidence: gradient updates concentrate probability on early successes (typically mediocre solutions), collapsing diversity. ``Towards Execution-Grounded Automated AI Research''~\cite{execution_grounded_ai2026} notes that RL from execution rewards can collapse to narrow ideas, exactly the over-sharpening we observe. Our results extend this analysis: in dense execution-grounded optimization, the best solutions occupy the low-probability tail of the model's distribution, so sharpening toward high-confidence modes moves in the wrong direction. The compute-optimal strategy is sampling for diversity plus intelligent selection, not further sharpening.

\textbf{Selection Strategies and Model Confidence.} Prior work on selection typically favors highest-confidence outputs or uses reward models for reranking. Snell et al.\ \cite{snell2024} establish that compute-optimal test-time strategies outperform naive Best-of-N through intelligent selection. Khatri et al.\ \cite{scaling_rl_compute2025} characterize how to optimally scale RL compute for LLMs, providing a framework for understanding compute allocation trade-offs. S*~\cite{sstar2025} achieves state-of-the-art code test-time scaling through Adaptive Input Synthesis (generating new test cases to differentiate candidates), but this requires additional LLM calls. For kernel optimization, we find that \textbf{surprisal-guided selection} (highest-surprisal, i.e., lowest log-probability, among correct samples) outperforms standard approaches by 30 percentage points with zero additional compute. The model's probability distribution already encodes this signal in the high-surprisal tail. This inverse relationship between confidence and quality has precedent in calibration literature. Guo et al.\ \cite{guo2017} show modern neural networks are often miscalibrated, but this has not been operationalized as a selection strategy for execution-grounded code generation.

\textbf{Verifiable Execution-Grounded Tasks.} We focus on VEG tasks: domains where a deterministic evaluator provides ground-truth feedback without human judgment. GPU kernel optimization is the primary example: KernelBench~\cite{kernelbench2025} evaluates 250 workloads on correctness and speedup, with speedup ranging continuously from 0x to 10x+. Related VEG domains include assembly superoptimization~\cite{supercoder2025} and formal theorem proving. The VEG setting enables our surprisal-guided selection strategy: execution feedback allows filtering to correct samples before applying surprisal-based selection.

\textbf{Kernel Optimization.} Prior work on LLM-based kernel optimization has not studied selection strategies. Kevin~\cite{kevin2025} achieves 82\% correctness through multi-turn train-time RL but keeps weights frozen at inference. CUDA-L2~\cite{cudal2_2025} surpasses cuBLAS by 19.2\% through two-stage GRPO. Magellan~\cite{magellan2026} requires $\sim$1.5 days of evolutionary search. AccelOpt~\cite{accelopt2025} uses ``Optimization Memory'' for kernel search. Concurrent work develops richer training-time RL pipelines: Dr.\ Kernel~\cite{drkernel2026} introduces KernelGYM with reward hacking checks and TRLOO, a debiased alternative to GRPO, achieving 31.6\% Fast@1.2 on KernelBench L2 through sequential multi-turn refinement; CUDA-L1~\cite{cudal1_2025} trains a contrastive RL model achieving 3.12$\times$ average speedup across all 250 KernelBench tasks; QiMeng-Kernel~\cite{qimeng2025} decouples strategy from implementation via hierarchical RL. These works optimize how to \emph{train} kernel-generating policies. We study a different question: given a capable policy, how should \emph{test-time compute} be allocated between sampling, selection, and gradient adaptation?

\section{Method}

\subsection{Dual-Loop Architecture}

To answer ``is adaptation the right strategy for dense-reward VEG tasks?'', we require controlled comparison between gradient-based adaptation and pure search strategies. We address this through a \textbf{dual-loop architecture} (Figure~\ref{fig:architecture}):

\textbf{Outer Loop (Train-Time RLVR).} The outer loop uses reinforcement learning from verifiable rewards (RLVR) to establish generalization across tasks. We train a base policy on 80 KernelBench L1 training tasks using Group Relative Policy Optimization (GRPO)~\cite{grpo2024} with LoRA adaptation. This produces a capable checkpoint (98.4\% correctness, 0.87x mean speedup) that serves as the shared starting point for all test-time strategies.

\textbf{Inner Loop (Test-Time Strategies).} The inner loop is the experimental arena. Given the trained base policy and a held-out evaluation set (5 tasks), we compare three test-time strategies under matched compute budgets: \textbf{Best-of-N search}, which samples $K$ candidates and selects via oracle, surprisal-guided, or random selection; \textbf{batch TTT adaptation}, which performs gradient updates with Best-of-Adaptation (BoA) checkpoint selection; and \textbf{Self-Distilled Policy Optimization (SDPO)}, which applies token-level distillation with or without execution feedback.

All strategies use the same base checkpoint, temperature (0.25), and maximum tokens (1024). The key variable is \emph{how} test-time compute is allocated: sampling diversity vs.\ gradient adaptation.

\textbf{Shared Evaluator.} Both loops use the identical KernelBench execution-grounded evaluator (Section~\ref{sec:environment}). This ensures that ``correct'' and ``speedup'' mean the same thing across all strategies, eliminating confounds from evaluation protocol differences.

\subsection{Execution-Grounded Environment}
\label{sec:environment}

Both train-time and test-time phases share KernelBench's deterministic evaluator, which provides dense scalar rewards through a five-stage pipeline. Model output is first inserted into a scaffolded kernel template, then compiled with CUDA error capture. Correct samples are those that pass functional equivalence tests against the reference implementation. For correct samples, timing is measured via CUDA events with median taken over multiple trials. The reward is computed as $\text{speedup} = \text{baseline\_time} / \text{kernel\_time}$, with incorrect samples receiving zero reward.

The continuous nature of the speedup reward (ranging from 0x for incorrect to 10x+ for highly optimized kernels) distinguishes this domain from preference-based tasks where rewards are binary or sparse. Every sample provides gradient signal proportional to its quality, enabling efficient gradient aggregation across diverse rollouts.

\subsection{Train-Time: RLVR (Outer Loop)}

Training uses normalized rewards (baseline-relative speedup per task) for stability across tasks with varying baseline performance. See Appendix~\ref{app:config} for the full outer loop specification.

\subsection{Test-Time Strategies (Inner Loop)}

At test time, we adapt the trained checkpoint to held-out evaluation tasks using batch updates inspired by Test-Time Reinforcement Learning (TTRL)~\cite{ttrl2025}. Each adaptation step processes $N\!=\!5$ tasks jointly, sampling $K\!=\!32$ rollouts per task to produce 160 samples per step. A GRPO gradient update is computed across all rollouts, and the rank-16 LoRA adapter is updated in-place. Unlike TTT-Discover's $\sim$50-step full RL, we use minimal adaptation (1--5 steps) to enable controlled comparison with search baselines under matched sample budgets.

\subsection{Best-of-Adaptation (BoA)}

We define \textbf{Best-of-Adaptation (BoA)} as checkpoint selection over an adaptation trajectory. Instead of assuming the final checkpoint is best, BoA selects $\arg\max(\text{fast\_1})$ across all steps.

\begin{algorithm}[t]
\caption{BoA with In-Batch Validation}
\label{alg:boa}
\begin{algorithmic}[1]
\REQUIRE Tasks $T$, checkpoint $\theta_0$, steps $S$, rollouts $K$
\STATE $\text{scores}[0] \gets \text{evaluate}(\theta_0, T)$
\FOR{$s = 1$ to $S$}
    \STATE $\text{rollouts} \gets \text{sample}(\theta_{s-1}, T, K)$
    \STATE $\theta_s \gets \text{gradient\_update}(\theta_{s-1}, \text{rollouts})$
    \STATE $\text{scores}[s] \gets \text{aggregate\_fast\_1}(\text{rollouts})$
\ENDFOR
\RETURN $\theta_{\arg\max(\text{scores})}$
\end{algorithmic}
\end{algorithm}

\textbf{Early Stopping Variant}: Stop when validation regresses for $P$ consecutive steps. In our experiments, $P\!=\!1$ matched oracle selection; the first regression signaled the optimal checkpoint.

\subsection{SDPO: Execution-Grounded Self-Distillation}

We extend batch adaptation with \textbf{Self-Distilled Policy Optimization (SDPO)}~\cite{sdpo2026}, replacing scalar reward advantages with token-level self-distillation signal conditioned on execution feedback. The teacher scores the student's sampled tokens:
\begin{equation}
A_t = \beta \cdot (\log p_{\text{teacher}}(x_t | \text{context}) - \log p_{\text{student}}(x_t | \text{prompt}))
\end{equation}
where $\beta = 1.0$ controls the distillation strength. Full SDPO methodology and results are presented in Appendix~\ref{app:sdpo}.

\section{Experimental Setup}

\subsection{Equal-Budget Protocol}

The key methodological contribution is rigorous budget matching. All parameters are held constant between Best-of-N and batch TTT: 320 total rollouts, temperature 0.25, max\_tokens 1024, fast evaluation mode, the final RLVR checkpoint, and the optimized system prompt. The only variable is \emph{how} the 320 rollouts are used: independently (Best-of-N) or as gradient signal (TTT).

\subsection{Baselines}

We evaluate against three baselines. \textbf{Best-of-N ($K\!=\!64$)} samples 64 candidates per task and selects the highest fast\_1, totaling 320 rollouts across 5 tasks. The \textbf{base policy} uses the RLVR checkpoint without test-time adaptation (step~0). \textbf{SDPO} variants (feedback and prompt-only self-distillation) are reported in Appendix~\ref{app:sdpo}.

\subsection{Metrics}

We report three metrics. \textbf{fast\_1} measures the fraction of samples that are both correct and achieve speedup $> 1$x; this is our primary metric as it captures both functional validity and performance improvement. \textbf{Correctness} measures the fraction passing functional equivalence tests. \textbf{Mean speedup} reports the average speedup across correct samples.

\subsection{Tasks}

\textbf{Subset 1 (Primary)}: Tasks \{4, 5, 12, 14, 15\} from KernelBench L1 eval split.

\textbf{Subset 2 (Robustness)}: Tasks \{18, 28, 29, 30, 32\} (offset=5).

\textbf{Extended Eval (Full L1)}: Tasks \{36, 55, 65, 70, 76, 82, 87, 89, 95, 98\} complete the full 20-task L1 eval set (seed 42). Best-of-N, per-sample evaluation, and logprob analysis reported; selection strategy analysis in Section~\ref{sec:surprisal}.

\subsection{Selection Strategies}

We compare five selection strategies for choosing among $K\!=\!64$ samples per task:

\begin{table}[tb]
\caption{Selection strategies for Best-of-N samples.}
\label{tab:strategies}
\centering
\footnotesize
\begin{tabular}{@{}lll@{}}
\toprule
Strategy & Selection Rule & Cost \\
\midrule
best-correct (Oracle) & Max speedup, correct only & Post-hoc \\
random-correct & Random correct sample & Baseline \\
confidence-guided & Max logprob, correct only & Zero \\
\textbf{surprisal-guided} & \textbf{Min logprob, correct only} & \textbf{Zero} \\
\textbf{surprisal-guided-top3} & \textbf{Best of 3 min-logprob} & \textbf{3 evals} \\
\bottomrule
\end{tabular}
\parbox{\columnwidth}{\scriptsize\textit{All strategies share correctness-checking of all $K$ samples. The ``Cost'' column shows additional speedup timing evaluations beyond shared filtering.}}
\end{table}

The surprisal-guided strategies are motivated by probing experiments (Section~\ref{sec:surprisal}) showing an inverse correlation between model confidence and kernel quality. The intuition: the model's highest-surprisal correct samples are solutions it ``didn't expect to find,'' often the creative, hardware-aware optimizations that yield maximum speedup.

\subsection{Compute Accounting}

We report \textbf{rollouts and total tokens (student + teacher)} for each method. See Appendix~\ref{app:compute} for full token accounting.

\section{Results}

\subsection{Main Result: Search Outperforms Minimal Adaptation}

\begin{figure}[t]
\centering
\includegraphics[width=\columnwidth]{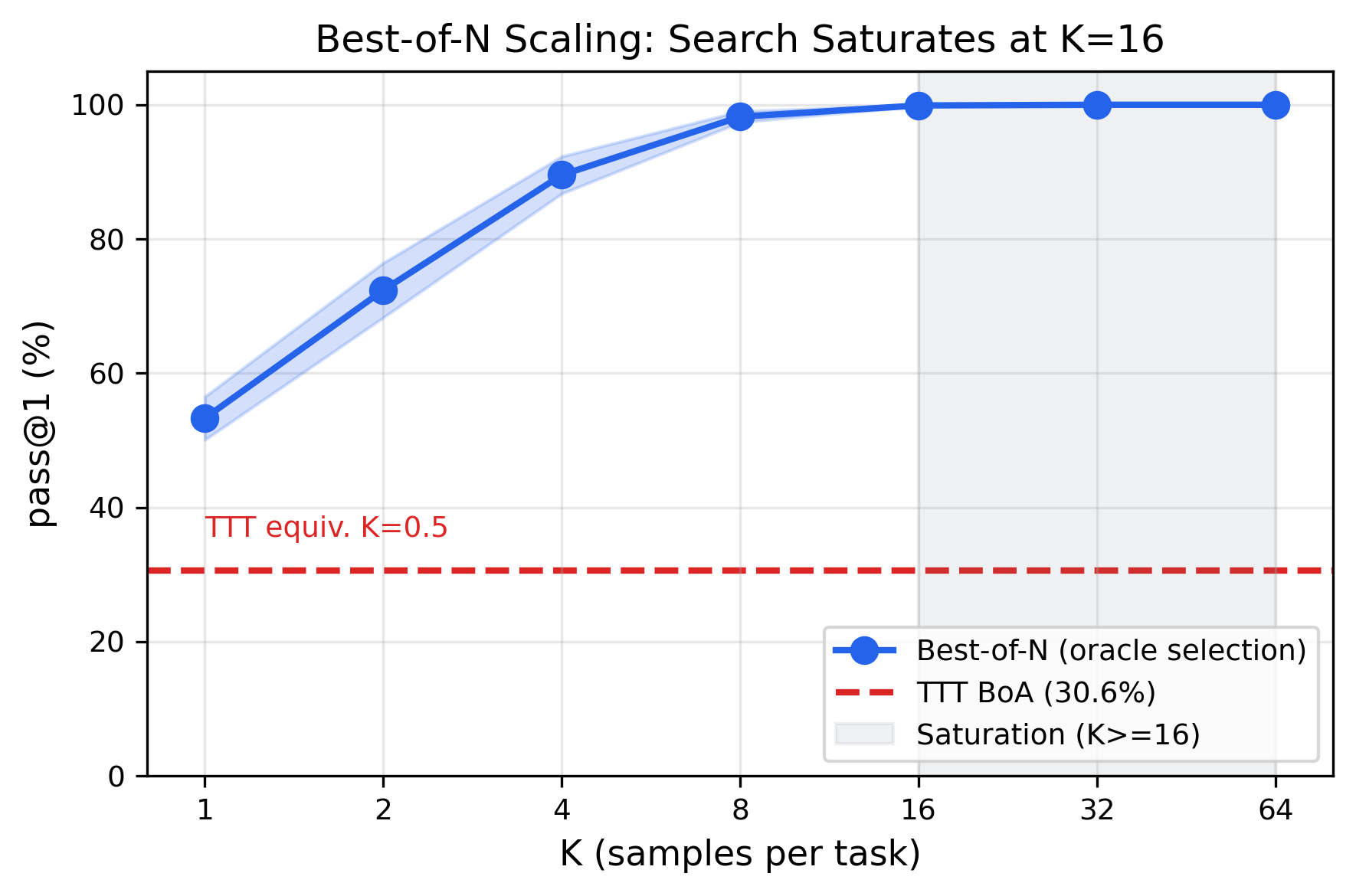}
\caption{\textbf{Best-of-N scaling curve.} Performance saturates at $K\!=\!16$ (99.9\%). TTT BoA at 30.6\% falls below $K\!=\!1$ random sampling (53.3\%).}
\label{fig:scaling}
\end{figure}

The central finding is that Best-of-N search decisively outperforms test-time training under matched compute budgets. This section presents the scaling curve comparison (Figure~\ref{fig:scaling}); Section~\ref{sec:surprisal} presents our surprisal-guided selection strategy that achieves oracle-matching performance.

\begin{table}[t]
\caption{Best-of-N scaling (Subset 1, 2 seeds).}
\label{tab:scaling}
\centering
\small
\begin{tabular}{@{}lccc@{}}
\toprule
$K$ & pass@1 & std & 95\% CI \\
\midrule
1 & 53.3\% & 3.2\% & [20\%, 100\%] \\
2 & 72.4\% & 4.1\% & [40\%, 100\%] \\
4 & 89.5\% & 2.8\% & [60\%, 100\%] \\
8 & 98.2\% & 0.8\% & [80\%, 100\%] \\
16 & 99.9\% & 0.1\% & [100\%, 100\%] \\
32 & 100\% & 0\% & [100\%, 100\%] \\
64 & 100\% & 0\% & [100\%, 100\%] \\
\bottomrule
\end{tabular}
\vspace{0.3em}
\parbox{\columnwidth}{\scriptsize\textit{pass@1 denotes the probability that a randomly drawn sample is both correct and achieves speedup $> 1\times$ (i.e., fast\_1 as defined in Section~4.3).}}
\end{table}

Performance saturates at $K\!=\!16$ with 99.9\% success. Beyond this point, marginal gains are near-zero, establishing that modest sampling budgets suffice for dense-reward VEG tasks.

\textbf{TTT Equivalent $K < 1$}: TTT's Best-of-Adaptation achieves 30.6\% fast\_1 (3-seed mean). Interpolating on the scaling curve, this falls \emph{below} $K\!=\!1$ (53.3\%), meaning TTT is worse than drawing a single sample ($K\!=\!1$). Best-of-N at $K\!=\!64$ achieves 100\% (oracle upper bound).

\textbf{Why TTT Fails: Over-Sharpening.} The failure mode is distribution collapse. TTT gradient updates concentrate probability mass on solutions that succeeded early in adaptation, typically mediocre solutions that happened to work. This destroys the diversity needed to find optimal kernels, which lie in the low-probability tail of the distribution. Evidence for over-sharpening includes: (1) non-monotonic trajectory with performance peaking at step 1--2 then regressing (Section~\ref{sec:trajectory}), (2) high variance across seeds (TTT BoA std $= 11.3\%$ vs.\ Best-of-N std $= 3.2\%$), (3) task-specific collapse with no consistent optimum across tasks, and (4) direct negative log-likelihood (NLL) vs.\ speedup probe confirming progressive anti-calibration (Section~\ref{sec:probing}).

\subsection{Surprisal-Guided Selection}
\label{sec:surprisal}

\begin{figure}[t]
\centering
\includegraphics[width=\columnwidth]{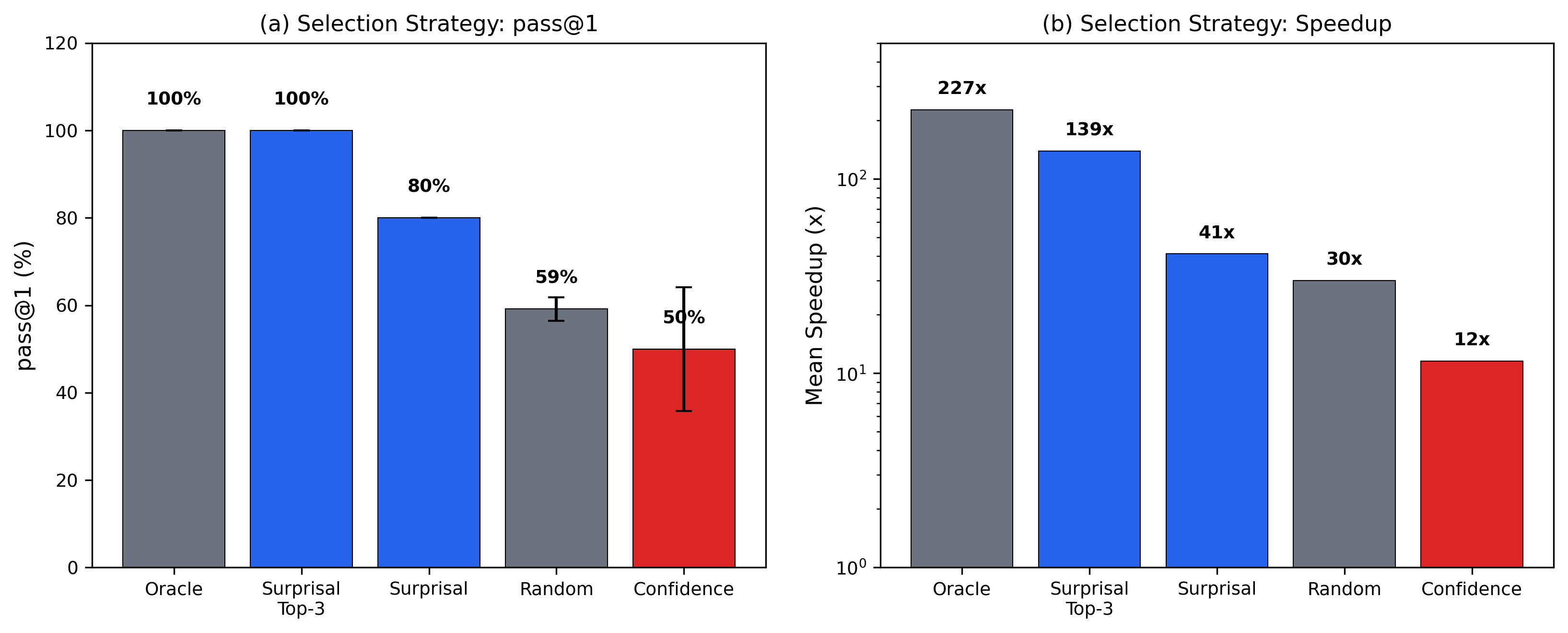}
\caption{\textbf{Selection strategy comparison.} (a) fast\_1 success rate and (b) mean speedup. Surprisal-guided achieves 80\% vs.\ 50\% for confidence-guided (+30\%). Surprisal-guided-top3 matches oracle.}
\label{fig:selection}
\end{figure}

Given that search outperforms minimal adaptation, how should we select among correct samples? Standard practice is to choose the highest-confidence (lowest-surprisal) output. We discover that for kernel optimization, the opposite strategy works better (Figure~\ref{fig:selection}). \emph{Critical distinction:} we select the most surprising \emph{correct} sample, not the most surprising overall (which would be gibberish). The VEG setting enables this by providing a correctness filter via execution.

\begin{table}[t]
\caption{Selection strategy results (Subset 1, 2 seeds).}
\label{tab:selection}
\centering
\small
\begin{tabular}{@{}lccc@{}}
\toprule
Strategy & fast\_1 & std & Mean Speedup \\
\midrule
best-correct (Oracle) & 100\% & 0\% & 226.9x \\
\textbf{surprisal-guided-top3} & \textbf{100\%} & \textbf{0\%} & \textbf{139.0x} \\
\textbf{surprisal-guided} & \textbf{80\%} & \textbf{0\%} & \textbf{41.2x} \\
random-correct & 59.2\% & 2.7\% & 30.0x \\
confidence-guided & 50\% & 14.1\% & 11.6x \\
\bottomrule
\end{tabular}
\end{table}

Three findings emerge from this comparison. First, surprisal-guided selection beats confidence-guided by 30\% (80\% vs.\ 50\%): selecting the \emph{highest-surprisal} correct sample outperforms the standard confidence-guided approach. Second, surprisal-guided-top3 matches oracle performance: evaluating just 3 samples (the 3 highest-surprisal correct ones, after the shared correctness filter) and selecting the fastest achieves 100\% success, identical to oracle selection over all 64 samples. Third, confidence-guided selection exhibits high variance (std $= 14.1\%$), indicating unreliable performance, while surprisal-guided has std $= 0\%$, consistent success across all tasks.

\textbf{Statistical strength.} The 30-percentage-point gap (80\% vs.\ 50\%) corresponds to Cohen's $h = 0.64$ (medium-to-large effect). On continuous speedup ratios, a paired Wilcoxon test on $\log(\text{speedup}_{\text{surprisal}} / \text{speedup}_{\text{confidence}})$ across 10 task-seed pairs yields $p = 0.084$ (test statistic $= 10.0$). On binary outcomes, all 3 discordant pairs favor surprisal (exact sign test $p = 0.125$, one-sided). Statistical power is limited at $n = 10$; Section~\ref{sec:limitations} discusses this. The logprob variance analysis below characterizes when the selection signal is available across the full 20-task eval.

\textbf{Why Does Surprisal-Guided Selection Work?} The model's probability distribution is a map of \emph{frequency}, not \emph{quality}. Because naive code is more common than expert-optimized code in training data, the model's ``confidence'' (log-probability) is a proxy for how \emph{common} a strategy is, not how \emph{fast} it is. By selecting for surprisal, we are explicitly filtering for what we call the \textbf{Expert Tail}: those rare, high-performance strategies that the model knows how to generate but considers statistically unlikely compared to naive idioms.

Concretely, high-quality kernels often require unusual memory access patterns, creative loop structures that deviate from common idioms, and hardware-specific optimizations not well-represented in training data. These rare solutions occupy high-surprisal regions of the model's distribution. Unlike S*~\cite{sstar2025}, which requires additional LLM calls to differentiate candidates, surprisal-guided selection extracts this signal at zero cost from existing log-probabilities.

\textbf{Length-Controlled Analysis.} A potential confound: longer code might have lower log-probability simply due to accumulating more token probabilities. Three analyses control for this. The raw Spearman correlation between logprob and speedup is weak ($\rho = -0.047$, $p = 0.27$). The partial correlation controlling for code length is essentially zero ($\rho = 0.003$, $p = 0.95$). And the direct length-speedup correlation is negligible ($\rho = -0.039$). The surprisal-guided effect is not explained by code length.

\textbf{Why weak correlation coexists with strong selection.} The near-zero correlation may appear to contradict the 80\% vs.\ 50\% selection result, but these measure fundamentally different things. Correlation measures whether surprisal \emph{linearly predicts} speedup across all 550 correct samples, and it does not. Selection operates via \emph{per-task argmax}: for each task, we pick the single highest-surprisal correct sample. \emph{Selection operates on the per-task argmax in the tail, not on the global slope.} This succeeds when the highest-surprisal sample within each task tends to be among its best solutions, a per-task ordinal property that global linear correlation cannot capture.

\textbf{Quartile Analysis.} The full quartile breakdown appears in Appendix~\ref{app:supplementary_tables} (Table~\ref{tab:quartiles}). Q2 (second-highest surprisal) shows the highest fast\_1 (81.0\%), while Q4 (lowest surprisal) shows the lowest (43.9\%). The high-surprisal half (Q1+Q2) averages 64.2\% fast\_1 versus 58.1\% for the low-surprisal half (Q3+Q4). This pattern suggests the optimal selection point is in the high-surprisal region but not the extreme tail.

\textbf{Logprob Variance and Selection Applicability.} Expanding logprob analysis to all 20 L1 eval tasks reveals a prerequisite for surprisal-guided selection: sufficient intra-task logprob variance. Of 20 tasks, 9 exhibit high variance (logprob std $> 1.0$)---these include the original Subset~1 and~2 tasks, where diverse solution strategies create a logprob gradient for selection to exploit. The remaining 11 tasks produce near-identical logprobs across samples (std $< 0.15$ for 7 tasks). These are primarily convolution and normalization operations where the reference uses cuDNN (already near-optimal); the model converges to a narrow template with minimal speedup headroom (sample\_fast\_1 of 7--23\%, selected speedup 1.00--1.22x). On such tasks, all selection strategies degenerate to random. Head-to-head across the 19 valid tasks (excluding Task~95, 0\% correctness), surprisal wins 6/8 on high-variance tasks but only 3/11 on low-variance tasks. The surprisal-guided effect is concentrated in tasks where diverse optimization strategies exist---precisely the tasks where selection provides leverage.

\subsection{TTT Trajectory: Why Adaptation Fails}
\label{sec:trajectory}

\begin{figure}[t]
\centering
\includegraphics[width=\columnwidth]{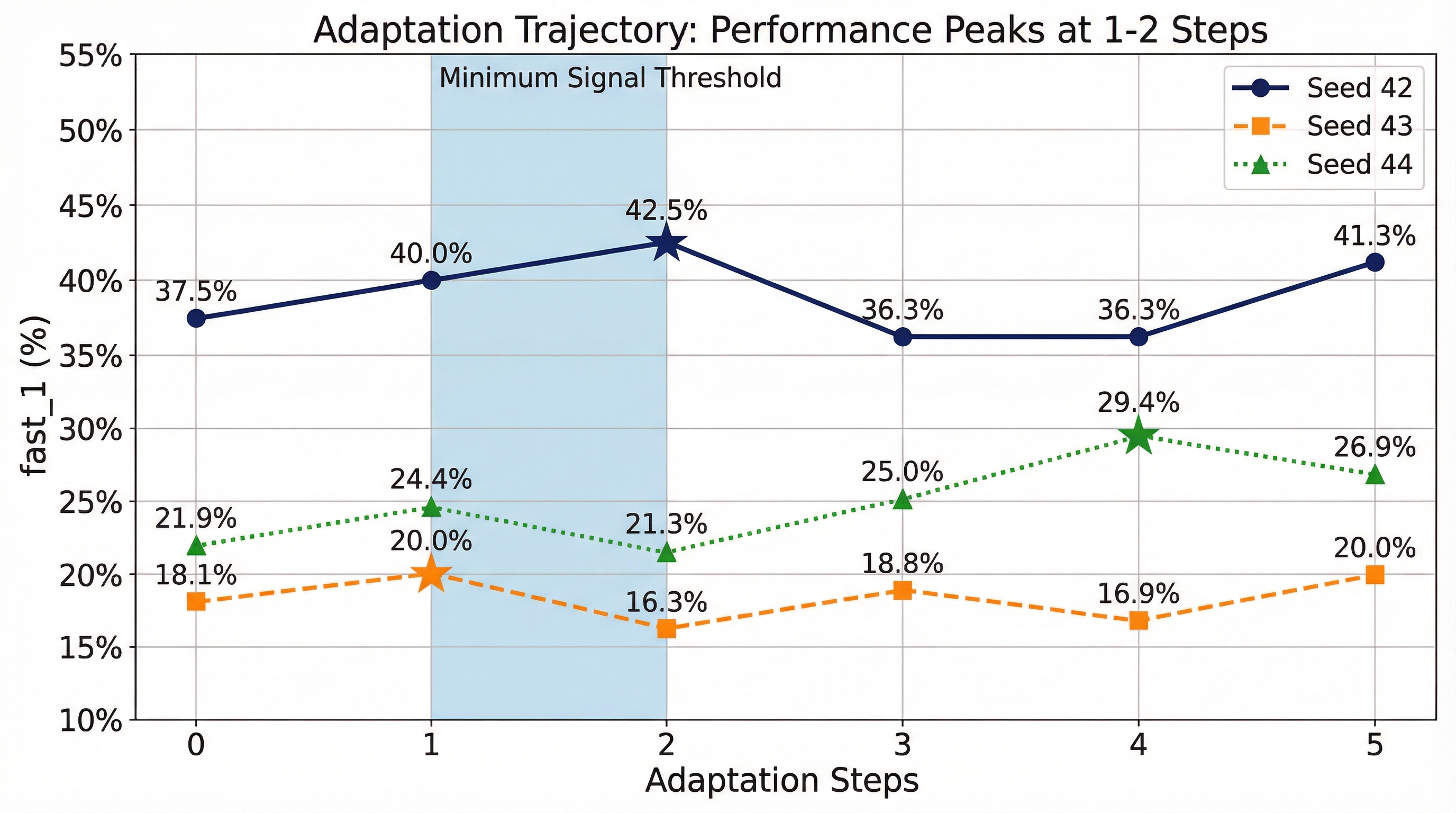}
\caption{\textbf{Adaptation trajectory.} Performance peaks at 1--2 steps then regresses. Stars mark BoA-selected checkpoints.}
\label{fig:trajectory}
\end{figure}

To understand why TTT underperforms, we examine the adaptation trajectory (Figure~\ref{fig:trajectory}). Performance peaks at step~2 (42.5\%) and regresses to 36.3\% at step~3, the over-sharpening dynamic in action (full per-task breakdown in Appendix~\ref{app:trajectory}). Gradient updates collapse diversity toward mediocre solutions.

The per-task breakdown reveals heterogeneous dynamics: Task~12 is already saturated at 100\% (easy task, no room for improvement); Task~5 peaks at step~1, then collapses to 3.1\% by step~3; Task~4 peaks at step~2, then regresses. This heterogeneity explains why TTT's aggregate BoA performance (30.6\% 3-seed mean) underperforms Best-of-N (100\% at $K\!=\!16$): adaptation cannot simultaneously optimize for tasks with different optima.

\textbf{Per-Task Isolated TTT Validation.} To confirm that over-sharpening occurs even without multi-task interference, we ran TTT on individual tasks (batch\_size$=1$, $K\!=\!32$, 5 steps; full results in Appendix~\ref{app:supplementary_tables}, Table~\ref{tab:pertask}). The pattern reveals task-level heterogeneity: Tasks~4, 5, 15 show classic over-sharpening (peak at step~1, then regress); Task~14 shows a late peak at step~4; Task~12 is saturated throughout. This confirms the heterogeneity observed in the batch setting, ruling out multi-task interference as the sole cause of over-sharpening.

Across all per-task isolated TTT runs (Table~\ref{tab:pertask}), 4 of 5 tasks peak within the first 2 adaptation steps (Tasks~4, 5, 12, 15); only Task~14 shows a delayed peak at step~4.

\subsection{Probing Over-Sharpening: Direct Measurement}
\label{sec:probing}

The evidence for over-sharpening in Sections~\ref{sec:trajectory} and the per-task analysis above is outcomes-based: performance regresses after step 1--2. To directly measure the probability shift, we probe how the adapted policy's NLL rankings relate to sample quality across TTT steps.

We take 320 fixed Best-of-N samples ($K\!=\!64$, seed 42) with known speedups and compute each sample's NLL under every TTT checkpoint (steps 0--8). This holds the sample set constant while varying only the scoring model, isolating the effect of adaptation on the probability-quality relationship.

\begin{figure}[t]
\centering
\includegraphics[width=\columnwidth]{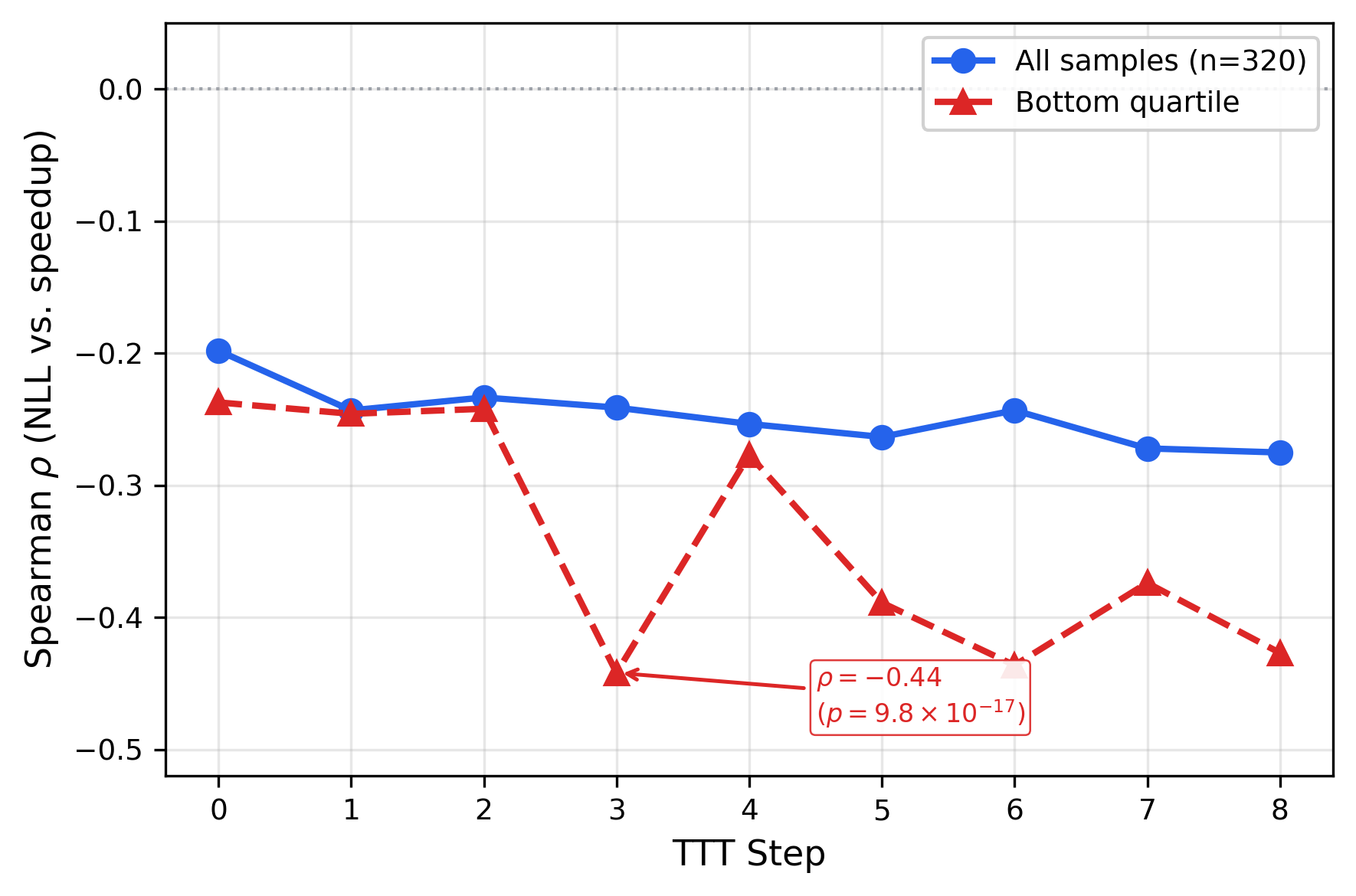}
\caption{\textbf{Over-sharpening probe.} Spearman $\rho$(NLL, speedup) across TTT steps for 320 fixed samples. The negative correlation deepens overall: adaptation makes the model progressively more confident about its worst solutions. Bottom-quartile $\rho$ nearly doubles from $-0.24$ to $-0.44$.}
\label{fig:rho_probe}
\end{figure}

Figure~\ref{fig:rho_probe} shows the result. The Spearman $\rho$ between NLL and speedup deepens overall from $-0.198$ (step~0, $p = 3.6 \times 10^{-4}$) to $-0.275$ (step~8, $p = 5.7 \times 10^{-7}$). The bottom-quartile correlation (the performance-critical tail where selection operates) nearly doubles from $\rho = -0.237$ to $\rho = -0.442$ (step~3, $p = 9.8 \times 10^{-17}$).

This is not merely diversity loss. The deepening negative $\rho$ means TTT assigns progressively higher confidence to worse solutions: active anti-calibration in exactly the region where surprisal-guided selection operates. Meanwhile, the mean NLL trajectory oscillates non-monotonically ($6.71 \to 6.66 \to 6.75 \to 6.86$), indicating unstable training dynamics rather than smooth convergence.

\subsection{Robustness: Second Task Subset (Hard Regime)}

To test whether findings generalize, we replicated the comparison on Subset~2: tasks \{18, 28, 29, 30, 32\} (offset=5 from the eval split), a harder 5-task subset.

\begin{table}[tb]
\caption{Subset 2 results (seed 42).}
\label{tab:subset2}
\centering
\small
\begin{tabular}{@{}lccc@{}}
\toprule
Method & Rollouts & Agg fast\_1 & Delta \\
\midrule
Best-of-N ($K\!=\!64$) & 320 & 36.9\% & baseline \\
BoA Step 0 & 160 & 17.5\% & $-$19.4\% \\
BoA Step 1 & 320 & 16.3\% & $-$20.6\% \\
\bottomrule
\end{tabular}
\end{table}

\begin{figure}[tb]
\centering
\includegraphics[width=\columnwidth]{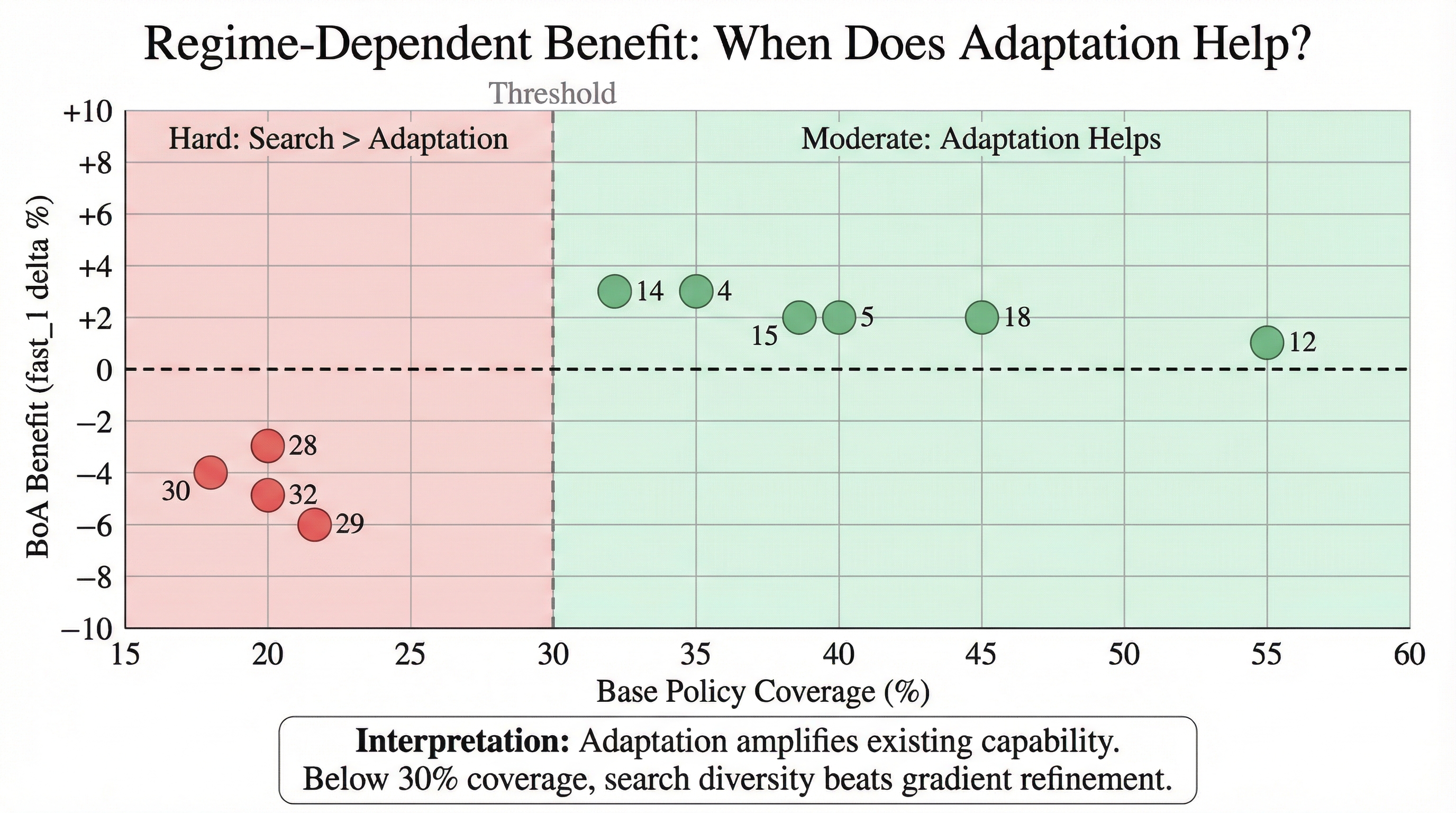}
\caption{\textbf{Regime-dependent benefit of adaptation vs.\ search.} Tasks with $>$30\% base coverage (light shading) show smaller BoA deficits; tasks below this threshold (dark shading) show larger deficits, favoring Best-of-N.}
\label{fig:regime}
\end{figure}

On the hard subset, Best-of-N outperforms BoA under equal budget. The contrasting results reveal that adaptation's deficit is regime-dependent: on Subset~1 (3 seeds), BoA shows a $-$9.2\% deficit relative to Best-of-N baseline (39.8\%), while on Subset~2 (seed 42), the deficit is larger at $-$20.6\% relative to baseline (36.9\%). Adaptation amplifies existing capability rather than creating new capability (Figure~\ref{fig:regime}). In both regimes, search maintains a diversity advantage.

\textbf{Full L1 Eval (20 Tasks).} Expanding Best-of-N to all 20 L1 eval tasks: 18/20 (90\%) achieve fast\_1$=1$ at $K\!=\!64$. Two failures are informative edge cases. Task~82 achieves 100\% correctness but 1.00x speedup---the reference uses cuDNN, leaving no optimization headroom. Task~95 achieves 0\% correctness---the model cannot generate valid kernels. Neither reflects a search strategy limitation: Task~82 is a task-level ceiling (no amount of search helps when the reference is already optimal); Task~95 is a model capability gap (no correct samples exist to select from). The 90\% success rate across the full eval set, versus 100\% on the 10-task subsets, reflects inclusion of tasks at both extremes of difficulty.

\textbf{Evaluation Note.} Results use a fast-proxy evaluation protocol (5 performance trials per kernel) rather than the full KernelBench benchmark (50 trials), as our focus is on test-time compute scaling strategies rather than benchmark leaderboard performance.

\subsection{Compression Mechanism: Rapid Signal Distillation}

We propose a \emph{compression view} of test-time training in VEG domains. The first 1--2 updates aggregate dense gradient signal from diverse rollouts, rapidly compressing execution feedback into the weights. Additional steps over-sharpen the distribution around a narrow subset of solutions, reducing diversity and causing regression.

The following observations support this interpretation. The BoA peak at 1--2 steps shows compression completing quickly. The step-dependent regression shows over-compression destroying diversity. SDPO experiments confirm this interpretation (Section~\ref{sec:sdpo_discussion}).

We hypothesize a \textbf{Reward Compression Principle}: \emph{gradient steps to saturation scale inversely with reward density.} Dense continuous rewards compress in 1--2 steps; sparse binary rewards require extended adaptation. This principle unifies our findings: fast saturation and feedback redundancy are two manifestations of the same underlying dynamic.

\textbf{Cross-Subset Transfer.} If adaptation builds generalizable optimization strategies, knowledge should transfer across task subsets. We evaluate cross-subset transfer: checkpoints adapted on Subset~1 (tasks \{4, 5, 12, 14, 15\}) evaluated on Subset~2 (tasks \{18, 28, 29, 30, 32\}) and vice versa.

Checkpoints adapted on Subset~1 and evaluated on Subset~2 achieve 7.5\% fast\_1, down from the unadapted baseline of 17.5\% and far below Best-of-N's 36.9\%. The reverse direction (Subset~2 adapted, Subset~1 evaluated) yields 31.25\%, also below the unadapted 37.5\%.

Both directions show degradation relative to unadapted baselines. Adaptation over-fits to training-subset modes rather than learning domain-general kernel optimization strategies, consistent with the over-sharpening interpretation.

\section{Discussion}

\subsection{Why VEG Tasks Favor Search Over Adaptation}

Verifiable execution-grounded tasks differ fundamentally from preference-based or sparse-reward domains. Three properties drive this difference. First, VEG tasks provide dense scalar rewards; speedup is continuous (0x to 10x+), not binary. Second, VEG evaluation is evaluation-bound: CUDA compilation, warmup runs, and performance timing create per-sample overhead. Third, when the world provides dense continuous feedback, an AI teacher interpreting that feedback becomes redundant (Section~\ref{sec:sdpo_discussion}).

These properties suggest VEG may be a distinct regime for test-time compute allocation. For domains with dense continuous rewards from a deterministic evaluator, sample diversity with intelligent selection (surprisal-guided) may be more efficient than gradient adaptation.

\subsection{The Minimum Signal Threshold}

We introduce a \emph{minimum signal threshold}: the gradient signal needed before over-sharpening degrades diversity. For dense-reward kernel optimization, this threshold is remarkably low: 1--2 steps from 160 diverse samples. Scalable Power Sampling~\cite{scalable_power_sampling2026} corroborates this (RL gains $=$ sharpening, not discovery), as does Agent RL Scaling Law~\cite{agent_rl_scaling2025} (models quickly internalize code heuristics). The threshold likely depends on reward density: sparse-reward domains may require more steps, while dense-reward domains saturate quickly.

\subsection{Relationship to Concurrent Work}
\label{sec:concurrent}

TTT-Discover~\cite{tttdiscover2026} represents the strongest case for extended test-time adaptation. Key differences distinguish our findings. First, TTT-Discover uses $\sim$50 steps while we find 1--2 optimal. Second, TTT-Discover does not compare against Best-of-N under matched compute. Third, the difference stems from reward density: sparse-reward discovery may require extended exploration, while dense-reward VEG tasks saturate quickly.

Our results complement rather than contradict TTT-Discover by identifying reward density as the key variable determining optimal adaptation duration.

Our TTT experiments use vanilla GRPO without the entropy regularization, reuse buffers, and extended rollout budgets (512 per step) employed by TTT-Discover~\cite{tttdiscover2026}. Our total budget is 320 rollouts versus TTT-Discover's 25,600. Whether those mechanisms and budgets prevent over-sharpening in dense-reward VEG domains remains untested. Dr.\ Kernel's TRLOO algorithm~\cite{drkernel2026} addresses GRPO's self-inclusion bias; whether TRLOO prevents over-sharpening in dense-reward VEG TTT remains an open question.

\textbf{Objective mismatch.} TTT-Discover optimizes for ``find a new SOTA solution'' and returns the best solution across all steps (argmax reward). Our evaluation uses fast\_1 (correct and speedup $> 1\times$) on KernelBench L1 with $K$ capped at 64. These are different targets; our findings apply to budgeted inference, not open-ended discovery.

We swept learning rates across three orders of magnitude (lr $\in \{$1e-5, 1e-6, 3e-7$\}$; Appendix~\ref{app:lr_ablation}). Over-sharpening persists at all learning rates: lr=1e-6 peaks at step~1 (55.0\%) then regresses; lr=3e-7 never exceeds the unadapted baseline (31.9\%).

\subsection{Future Direction: Zero-Evaluation Discovery}
\label{sec:world_models}

Our findings point toward zero-evaluation discovery: models generating optimal code for novel hardware without physical execution. This requires physics-grounded world models, internal simulations of how code interacts with hardware. From Word to World~\cite{from_word_to_world2025} formalizes evaluation of implicit world models; SSRL~\cite{ssrl2025} proposes LLMs as internal world simulators. Execution feedback can be efficiently distilled into weights through minimal adaptation, a tractable path toward internalized hardware models.

Implicit hardware world models remain a research direction, not a demonstrated capability of this work. Our contribution is the efficiency characterization that makes this direction tractable.

\subsection{Self-Distillation: Why Feedback Adds No Lift}
\label{sec:sdpo_discussion}

SDPO experiments (Appendix~\ref{app:sdpo}) show prompt-only self-distillation succeeds at 120B scale, but feedback context provides no lift in VEG domains. When the world provides continuous gradient signal, feedback interpretation becomes redundant.

\subsection{Open Questions}

Several open questions remain. First, an \emph{entropy-regularized TTT} baseline (closer to TTT-Discover's entropic objective) would test whether the over-sharpening failure is fundamental to dense-reward VEG domains or specific to vanilla GRPO. Second, a \emph{deployable surprisal-guided pipeline} that ranks candidates by surprisal \emph{before} correctness filtering (evaluating only the top-$m$ candidates end-to-end) would transform the selection insight from post-hoc analysis to a practical method. Third, \emph{reward sparsification} experiments (thresholding the continuous speedup signal) would test whether the step-optimum shifts upward as reward becomes sparser, directly validating the reward compression hypothesis.

\section{Limitations}
\label{sec:limitations}

This work has several limitations. Best-of-N evaluation covers all 20 KernelBench L1 eval tasks (90\% task success at $K\!=\!64$). The primary selection experiment uses 10 task-seed pairs (5 tasks $\times$ 2 seeds); the 20-task logprob expansion reveals the variance regime boundary but uses a single seed. All experiments use a single 120B-parameter model; transfer to other sizes and architectures remains untested. We evaluate only L1 tasks; harder L2 and L3 levels may exhibit different dynamics.

Our TTT experiments scope to vanilla GRPO (see Section~\ref{sec:concurrent} for discussion of differences from TTT-Discover).

With $n=10$ binary outcomes (5 tasks $\times$ 2 seeds), our primary selection comparison (80\% vs.\ 50\%) has limited statistical power for binary tests (exact sign test $p = 0.125$). We supplement with continuous speedup analysis (Section~\ref{sec:surprisal}).

The surprisal-guided selection strategy requires multiple correct samples per task; on harder levels (L2/L3) or tasks where the base policy has low coverage, the effect may diminish or vanish. Moreover, surprisal-guided selection requires sufficient intra-task logprob variance. On 11/20 L1 tasks where the model produces near-identical solutions (logprob std $< 1.0$), the selection signal vanishes and all strategies perform equivalently.

Our evaluation uses a fast-proxy protocol (5 timing trials per kernel); rankings could shift under the full KernelBench protocol (50 trials). We validated selected kernels on H100 hardware and observed consistent rankings, but a full-protocol replication on a larger subset would strengthen these results.

The inverse relationship between confidence and quality may be domain-specific. In kernel optimization, rare creative solutions yield high speedups. In domains where the mode represents optimal behavior, surprisal-guided selection could perform poorly. A formal analysis of the relationship between training distribution coverage and test-time solution quality could strengthen these empirical findings.

\section{Conclusion}

We study compute-optimal test-time strategies for verifiable execution-grounded tasks, demonstrating across all 20 KernelBench L1 eval tasks that search outperforms minimal adaptation (1-5 gradient steps) for GPU kernel optimization. Best-of-N sampling achieves 90\% task success (18/20) at $K\!=\!64$ while TTT's best checkpoint reaches only 30.6\% (3-seed mean), with TTT's ``equivalent $K$'' falling below 1, meaning adaptation underperforms single-sample inference.

Three findings characterize the compute-optimal strategy:

\vspace{0.5em}
\begin{enumerate}
\item \textbf{Search saturates at modest $K$.} Best-of-N scaling shows performance saturates at $K\!=\!16$ (99.9\% success). Practitioners should invest in sample diversity, not gradient updates.

\item \textbf{Surprisal-guided selection recovers oracle performance.} Selecting the highest-surprisal correct sample achieves 80\% success vs.\ 50\% for most-confident, a 30\% improvement with zero additional compute. Extending to surprisal-guided-top3 matches oracle at 100\%.

\item \textbf{TTT fails due to over-sharpening.} Gradient updates collapse the policy toward mediocre solutions, destroying the diversity needed to find optimal kernels in the distribution tail.
\end{enumerate}

The practical implication is clear: for dense-reward VEG tasks, allocate compute to sample diversity and intelligent selection rather than gradient adaptation. The surprisal-guided selection principle (that rare, high-quality solutions occupy high-surprisal regions) may generalize to other execution-grounded domains where the optimum lies in the distribution tail.

The prevailing assumption that test-time training provides universal benefits does not hold in this regime. In domains with dense continuous rewards and deterministic evaluation, gradient updates that worked for sparse-reward reasoning tasks become counterproductive when the goal is finding optimal solutions in a well-sampled distribution's tail.

\section*{Acknowledgments}

We thank Thinking Machines Lab for access to the Tinker training infrastructure.

\bibliography{references}
\bibliographystyle{preprint}

\appendix

\section{Experimental Configuration}
\label{app:config}

{\scriptsize
\begin{verbatim}
# RLVR Training (produces base checkpoint)
model: openai/gpt-oss-120b
algorithm: GRPO
lora_rank: 16
learning_rate: 1e-5
batch_size: 8
group_size: 8
training_tasks: 80 (KernelBench L1 train split)
temperature: 0.25
max_tokens: 1024
normalize_reward: true

# Test-Time Evaluation
checkpoint: RLVR final (step 40,
            98.4% correct, 0.87x speedup)
eval_tasks: {4, 5, 12, 14, 15} (subset 1),
            {18, 28, 29, 30, 32} (subset 2)

# Best-of-N
K: 64
total_rollouts: 320 (5 tasks x 64)

# Batch TTT (BoA)
K: 32 per task
tasks_per_step: 5
learning_rate: 1e-5
step_1_rollouts: 320 (5 tasks x 32 x 2)
\end{verbatim}
}

\section{RLVR Training Progression}

Table~\ref{tab:rlvr_progression} shows the outer loop training progression. The final checkpoint (step 40) achieves 98.4\% correctness with 0.87x mean speedup, establishing a strong base policy for test-time evaluation.

\begin{table}[tb]
\caption{RLVR training progression (outer loop).}
\label{tab:rlvr_progression}
\centering
\begin{tabular}{@{}lccc@{}}
\toprule
Checkpoint & Correctness & Speedup & Notes \\
\midrule
Step 10 & 90.6\% & 0.81x & Early training \\
Step 20 & 95.3\% & 0.87x & --- \\
Step 30 & 95.3\% & 0.86x & --- \\
Step 40 (final) & 98.4\% & 0.87x & \textbf{Used for all eval} \\
\bottomrule
\end{tabular}
\end{table}

\section{TTT Trajectory Details}
\label{app:trajectory}

Table~\ref{tab:trajectory} shows the full per-task breakdown of the batch TTT adaptation trajectory for seed~42 on Subset~1. The aggregate performance peaks at step~2 (42.5\%) before regressing, while individual tasks show heterogeneous dynamics: Task~12 saturates at 100\% throughout; Tasks~4, 5, 15 peak early then regress; Task~14 shows delayed improvement.

\begin{table}[tb]
\caption{Batch TTT trajectory (seed 42, Subset 1).}
\label{tab:trajectory}
\centering
\small
\begin{tabular}{@{}lccccccc@{}}
\toprule
Step & Rollouts & Agg & T4 & T5 & T12 & T14 & T15 \\
\midrule
0 & 160 & 37.5 & 9.4 & 3.1 & 100 & 37.5 & 37.5 \\
1 & 320 & 40.0 & 28.1 & 15.6 & 100 & 21.9 & 34.4 \\
\textbf{2} & \textbf{480} & \textbf{42.5} & \textbf{46.9} & 6.3 & 100 & 18.8 & 40.6 \\
3 & 640 & 36.3 & 25.0 & 3.1 & 100 & 21.9 & 31.3 \\
4 & 800 & 36.3 & 21.9 & 3.1 & 100 & 15.6 & 40.6 \\
5 & 960 & 41.3 & 25.0 & 9.4 & 100 & 40.6 & 31.3 \\
\bottomrule
\end{tabular}
\end{table}

\section{Compute Accounting}
\label{app:compute}

Table~\ref{tab:compute} provides the full token accounting for each method, enabling compute-matched comparisons.

\begin{table}[tb]
\caption{Full compute breakdown (mean across seeds).}
\label{tab:compute}
\centering
\footnotesize
\setlength{\tabcolsep}{3pt}
\begin{tabular}{@{}lccccc@{}}
\toprule
Method & Rolls & Student & Teacher & Total & Wall (min) \\
\midrule
RLVR Base ($K\!=\!1$) & 5 & 4.0K & 0 & 4.0K & --- \\
Best-of-N ($K\!=\!64$) & 320 & 313K & 0 & 313K & 42 \\
BoA Step 1 & 320 & 313K & 0 & 313K & $\sim$60$^\dagger$ \\
SDPO (feedback) & 320 & 314K & 338K & 652K & --- \\
SDPO (prompt-only) & 320 & 311K & 313K & 625K & --- \\
\bottomrule
\end{tabular}
\vspace{0.3em}
\parbox{\columnwidth}{\scriptsize $^\dagger$BoA Step 1 matches BoN rollout count but includes backprop overhead; full 9-step TTT trajectory takes $\sim$267 min for 1,440 rollouts.}
\end{table}

\section{Speedup Statistics}

Raw speedup magnitude varies widely across tasks, reflecting task-specific optimization headroom rather than method quality. fast\_1 (correct AND speedup $> 1$x) is the operative metric.

\begin{table}[tb]
\caption{Best-of-N selected speedup ($K\!=\!64$, Subset 1, Seed 42).}
\centering
\small
\begin{tabular}{@{}lc@{}}
\toprule
Task & Selected Speedup \\
\midrule
4 & 859.8x \\
5 & 191.7x \\
12 & 23.8x \\
14 & 15.6x \\
15 & 13.6x \\
\bottomrule
\end{tabular}
\end{table}

\section{SDPO Self-Distillation Experiments}
\label{app:sdpo}

We extend batch adaptation with \textbf{Self-Distilled Policy Optimization (SDPO)}, replacing scalar reward advantages with token-level self-distillation signal conditioned on execution feedback.

\textbf{Teacher Context Construction.} For each student rollout, we construct a teacher context containing: (1) the original task prompt, (2) a correct solution from the same batch if available, (3) structured execution feedback (compile status, correctness, speedup, runtime, error traces), and (4) the instruction: ``Correctly solve the original question.'' The student's original code is \textbf{not} included; only its execution outcome.

Table~\ref{tab:sdpo_results} reports SDPO results across all three seeds on Subset~1.

\begin{table}[tb]
\caption{SDPO results (Subset 1, 3 seeds).}
\label{tab:sdpo_results}
\centering
\small
\begin{tabular}{@{}lcccc@{}}
\toprule
Method & Seed 42 & Seed 43 & Seed 44 & Mean $\pm$ std \\
\midrule
SDPO (feedback) & 35.6\% & 18.8\% & 24.4\% & 26.3\% $\pm$ 8.6\% \\
SDPO (prompt) & 38.8\% & 23.8\% & 28.8\% & 30.4\% $\pm$ 7.6\% \\
\bottomrule
\end{tabular}
\end{table}

\textbf{Key Finding: Feedback provides no lift.} SDPO with full execution feedback (26.3\%) underperforms SDPO prompt-only (30.4\%), consistent across all 3 seeds. In our KernelBench L1 setting and SDPO-style feedback construction, execution feedback provides no lift over prompt-only self-distillation. This is consistent with a \textbf{reward density hypothesis}: when the world provides dense continuous rewards, an AI teacher interpreting that feedback may be redundant. Whether this generalizes beyond our setting (to smaller models, harder tasks (L2/L3), or alternative feedback formats) remains open.

\textbf{Self-Distillation at Frontier Scale.} Using a 120B-parameter model, we find that prompt-only self-distillation succeeds at frontier scale, but feedback context provides no lift in VEG domains with dense rewards. This may change with model scale: at 120B, the model likely already encodes sufficient knowledge of error traces, making explicit feedback redundant. Smaller models or harder tasks (L2/L3) might benefit more from structured feedback context.

\section{Learning Rate Sensitivity}
\label{app:lr_ablation}

Table~\ref{tab:lr_ablation} shows TTT performance across three learning rates spanning three orders of magnitude. Over-sharpening persists at all learning rates.

\begin{table}[ht!]
\caption{TTT BoA across learning rates (Subset 1, seed 42).}
\label{tab:lr_ablation}
\centering
\small
\begin{tabular}{@{}lccl@{}}
\toprule
LR & Peak Step & BoA fast\_1 & Pattern \\
\midrule
1e-5 (default) & 2 & 42.5\% & Peak then regress \\
1e-6 & 1 & 55.0\% & Early peak, regression \\
3e-7 & 0 & 31.9\% & No improvement \\
\bottomrule
\end{tabular}
\end{table}

\section{Supplementary Tables}
\label{app:supplementary_tables}

\begin{table}[ht!]
\caption{Surprisal quartile breakdown.}
\label{tab:quartiles}
\centering
\footnotesize
\begin{tabular}{@{}lccc@{}}
\toprule
Quartile & fast\_1 & Speedup & Tokens \\
\midrule
Q1 (high surprisal) & 47.4\% & 37.0x & 7 \\
Q2 & 81.0\% & 26.2x & 8 \\
Q3 & 72.3\% & 46.8x & 10 \\
Q4 (low surprisal) & 43.9\% & 30.6x & 15 \\
\bottomrule
\end{tabular}
\parbox{\columnwidth}{\small\textit{Note:} Global quartiles conflate task identity with surprisal rank (Q1 is 79\% Task~5; Q2 is 48\% Task~12). Surprisal-guided selection operates per-task, avoiding this confound.}
\end{table}

\begin{table}[ht!]
\caption{Per-task isolated TTT (seed 42).}
\label{tab:pertask}
\centering
\small
\begin{tabular}{@{}lccccl@{}}
\toprule
Task & Step 0 & Step 1 & Step 2 & Steps 3--5 & BoA \\
\midrule
4 (hard) & 0\% & \textbf{6.3\%} & 0\% & 0\% & Step 1 \\
5 (mod.) & 3.1\% & \textbf{12.5\%} & 12.5\% & 3.1\% & Step 1 \\
12 (ctrl) & \textbf{100\%} & 100\% & 100\% & 100\% & Step 0 \\
14 (mod.) & 37.5\% & 34.4\% & 37.5\% & 56.3\% & Step 4 \\
15 (mod.) & 43.8\% & \textbf{46.9\%} & 34.4\% & 40.6--46.9\% & Step 1 \\
\bottomrule
\end{tabular}
\end{table}

\end{document}